\documentclass[AMA,STIX1COL]{WileyNJD-v5}

\usepackage{booktabs}
\usepackage{tabularx}
\usepackage{array}
\usepackage{xcolor}
\usepackage{listings}
\usepackage{enumitem}
\usepackage{url}
\hypersetup{hidelinks}

\newcolumntype{L}[1]{>{\hsize=#1\hsize\raggedright\arraybackslash}X}
\newcolumntype{C}[1]{>{\hsize=#1\hsize\centering\arraybackslash}X}

\definecolor{kw}{HTML}{1565C0}
\definecolor{cmt}{HTML}{6A6A6A}
\lstdefinestyle{pycode}{%
  language=Python,
  basicstyle=\ttfamily\footnotesize,
  keywordstyle=\color{kw}\bfseries,
  commentstyle=\color{cmt}\itshape,
  numbers=none, showstringspaces=false, breaklines=true,
  columns=fullflexible, xleftmargin=0.6em,
  aboveskip=0.4em, belowskip=0.4em,
  frame=tb, framesep=0.4em, framerule=0.4pt,
  rulecolor=\color{black!30},
}

\newcommand{\pcode}[1]{\texttt{#1}}
\newcommand{\AEROS}{\textsc{AEROS}}

\articletype{Research Article}
\journal{Software: Practice and Experience}
\volume{00}
\copyyear{2026}
\startpage{1}
\usepackage{placeins}

\historydates{}
\doiheadtext{}
\makeatletter
\def\oddfoot@titlepage@info{\vbox{\hsize\textwidth{\noindent\rule{\textwidth}{.5pt}\vskip-40pt {\fontsize{7}{8}\selectfont {\it \@journal} \@copyyear;\@volume:\@FirstPg--\@LastPg \hfill wileyonlinelibrary.com/journal/spe \quad \foot@vertbar\quad  {\pagenumfont\@FirstPg}}\vskip10pt }}}
\def\evenfoot@titlepage@info{\oddfoot@titlepage@info}
\makeatother

\begin{document}

\title{Bounded-Fidelity Sim-as-Demo-Stage: Mocap Handoff for Governance Benchmarks}

\author[1]{Xue Qin}
\author[2]{Simin Luan}
\author[3]{Cong Yang}
\author[2]{Zhijun Li}

\authormark{QIN \textsc{et al.}}
\titlemark{Bounded-Fidelity Sim-as-Demo-Stage}

\address[1]{\orgdiv{School of Software},
  \orgname{Harbin Institute of Technology},
  \orgaddress{\city{Harbin}, \country{China}}}
\address[2]{\orgdiv{School of Computer Science and Technology},
  \orgname{Harbin Institute of Technology},
  \orgaddress{\city{Harbin}, \country{China}}}
\address[3]{\orgdiv{School of Future Science and Engineering},
  \orgname{Soochow University},
  \orgaddress{\city{Suzhou}, \country{China}}}

\corres{Cong Yang, School of Future Science and Engineering,
  Soochow University, Suzhou, China. \email{cong.yang@suda.edu.cn}\\
  Zhijun Li, School of Computer Science and Technology,
  Harbin Institute of Technology, Harbin, China.
  \email{lizhijun\_os@hit.edu.cn}}

\abstract[Abstract]{Sim-to-real research pursues physics fidelity
as a primary objective: simulators are judged by how closely they
reproduce real-world contact dynamics. For \emph{governance
benchmarking} of LLM-driven robots, where the simulator
demonstrates that an admission/policy/contract/audit pipeline
behaves correctly, contact fidelity at object handoffs (grasp,
carry, place) becomes a liability: contact-force integration
noise injects audit-chain divergence that is structurally
unrelated to the governance property under test. We propose
\emph{bounded-fidelity sim-as-demo-stage}, a design pattern that
suppresses contact physics within explicitly bracketed
\emph{handoff envelopes} while preserving full dynamics
elsewhere. The construction uses MuJoCo's mocap-body primitive
driven by a $220$-line Python adapter that the governance bridge
invokes via structured intents. We formalise audit-chain
stability as byte-equality of the hashed event log across replays
and identify two structural envelope properties that imply it.
Across $N=1000$ replays per posture, the mocap variant produces
\emph{one} distinct audit-chain hash (1000/1000 byte-identical;
Wilson 95\% CI $[0.997, 1.000]$); the contact-force baseline
produces $584$ distinct hashes ($993/1000$ diverged; CI
$[0.987, 0.998]$). A timestep sweep ($\{1, 2, 5, 10\}$~ms) shows
the divergence is structural, not a tuning artefact: it stays at
$0.985$ at every timestep. Envelope-edge timing jitter ($\pm 10$
simulation steps, $1{,}400$ replays) produces $0$ divergence, and
audit chains remain byte-equal across $K \in \{1, 2, 3\}$
sequentially handed-off objects ($1{,}500$ replays) with
sub-linear per-pick-and-place overhead. The pattern gives
benchmark designers audit-chain reproducibility at near-zero
engineering cost; we also map where it is harmful (sim-to-real
validation, policy training, contact-rich tasks) so it is not
mis-deployed.}


\jnlcitation{\cname{%
\author{X.~Qin},
\author{S.~Luan},
\author{C.~Yang}, and
\author{Z.~Li}}.
\ctitle{Bounded-Fidelity Sim-as-Demo-Stage: Mocap Handoff for
Governance Benchmarks.}
\cjournal{\it Softw.\ Pract.\ Exper.} \cvol{2026;00(00):1--11}.}

\maketitle

\section{Introduction}\label{sec:intro}

A senior engineer evaluates a startup's robotic governance demo.
The pitch: an LLM-driven Franka Panda arm picks up a red cup,
carries it across a worktable, and places it in a drop-off tray.
The governance bridge admits the natural-language command, runs
the admission, policy, contract, and audit gates, and dispatches
the resulting structured intents to the simulator. The audit log
streams in real time alongside the 3D viewport.

Run~1 of the demo is clean. The arm reaches the cup, grasps it,
carries it, and releases it on the tray. The audit chain logs four
skill executions: \pcode{perception.locate\_object},
\pcode{navigation.move\_to}, \pcode{manipulation.grasp},
\pcode{manipulation.place}. Every gate shows admission GRANTED,
policy PASS, contract PASS, executor dispatched.

Run~2 of the same demo, same script, same seed. The arm reaches
the cup. The gripper closes. But this time the contact-force
integrator generates a brief spurious $87$~N spike from the
gripper-on-cup collision, the cup slips from the gripper's grasp,
and a moment later the cup geom clips the table edge and triggers
a runtime force-limit violation. The audit chain now records a
\pcode{recovery\_action} event the demo's narration does not
mention. The engineer is confused. Was that the bridge's bug, or
the simulator's?

This is the kernel of our argument. The senior engineer's
evaluation is about \emph{governance behaviour}, not about
\emph{physics fidelity}. The contact-force noise that spoiled
Run~2 was the high-fidelity contact integrator working as
intended. Precisely because it worked as intended, it leaked into
the audit chain and confused the evaluator about what they were
watching.

We propose \emph{bounded-fidelity sim-as-demo-stage} as a design
pattern for governance benchmarks: a simulator deliberately
configured to be deterministic at handoff points, even at the cost
of physics realism. The construction uses MuJoCo's
mocap-body primitive\citep{mujoco-todorov2012,mujoco-docs} to
implement object handoff via Python pose-tracking, sidestepping
the contact-force integrator during the carry phase. Across
reproducible replays of the construction, the audit chain is
byte-stable; under the contact-force equivalent it diverges.

\paragraph{Contributions.} (1)~The \emph{design choice} to build
governance-benchmark simulators around handoff suppression rather
than handoff fidelity, with an explicit problem-decomposition
argument for why the sim-to-real default is the wrong default in
this sub-context. (2)~A concrete implementation pattern using
MuJoCo mocap bodies driven by a $220$-line Python adapter
bracketed by governance-bridge structured intents, including a
formal definition of the \emph{handoff envelope} (Section~\ref{sec:formal})
with two structural properties (determinism preservation and
audit-chain stability under handoff) that admit lightweight
Hoare-style verification. (3)~An evaluation discipline that
treats audit-chain hash stability as a first-class metric, with
$N=1000$-replay byte-equality measurements, a four-point
timestep sweep ($200$ trials each) demonstrating that the
contact-force divergence is structural in the contact solver
rather than a tuning artefact, and two robustness ablations:
envelope-edge timing jitter ($\pm 10$ simulation steps over
$1{,}400$ replays) and compositional generalisation across
$K \in \{1, 2, 3\}$ sequentially handed-off objects (over
$1{,}500$ replays). (4)~An applicability matrix
(Section~\ref{sec:applicability}) identifying both the use cases
the pattern serves well (governance benchmarks, investor demos,
regression tests, tutorials) and the use cases for which it is
actively harmful (sim-to-real validation, policy training,
contact-rich tasks).

\paragraph{Roadmap.}
Section~\ref{sec:related} positions the work against sim-to-real,
LLM-driven robotics, governance-benchmark, reproducibility, and
kinematic-vs-dynamic-simulation literature. Section~\ref{sec:pattern}
presents the bounded-fidelity construction, including the
mocap-body mechanism, the handoff-envelope decomposition, and what
the pattern is and is not. Section~\ref{sec:formal} provides a
formal account: a state-transition definition of the envelope, a
specification of audit-chain stability as byte-equality, and two
structural properties stated in Hoare-style pre- and
post-conditions. Section~\ref{sec:eval} reports the implementation
and the $1000$-replay empirical study. Section~\ref{sec:discussion}
discusses the applicability boundary, including the explicit
when-not-to-use cases. Section~\ref{sec:conclusion} concludes.

\begin{figure}[t]
\centering
\includegraphics[width=\textwidth]{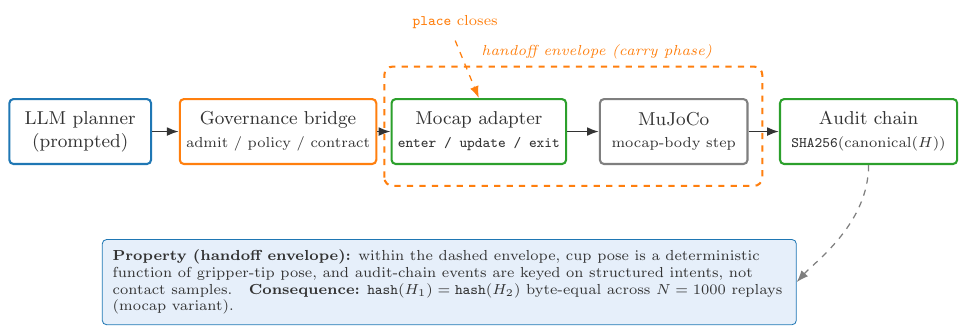}
\caption{System overview of the bounded-fidelity sim-as-demo-stage
pattern. The governance bridge admits a structured grasp intent,
dispatches it to the mocap adapter, and opens the handoff envelope
(orange dashed region) around the carry phase. Within the envelope,
the cup pose is written as a deterministic function of the gripper
pose, so the audit-chain hash is byte-equal across replays
(Proposition~\ref{prop:stability}; $1000/1000$ mocap variant vs
$7/1000$ contact-force baseline in \S\ref{sec:eval-quant}).}
\label{fig:hero}
\end{figure}

\section{Related Work}\label{sec:related}

\paragraph{Software design patterns.} The contribution is framed
as a software design pattern in the tradition of
Gamma et al.\citep{gamma1995gof} and Schmidt et al.\citep{schmidt2000posa2}: a named,
reusable solution to a recurring design problem, with explicit
applicability conditions and known consequences. The Bounded
Fidelity Pattern formalised in Section~\ref{sec:pattern} differs
from those classical pattern catalogues, which describe
compositional building blocks for object-oriented or concurrent
systems, by addressing a runtime-fidelity choice rather than a
class-structure choice. It inherits the same expository style of
context, problem, solution, and consequences.

\paragraph{Sim-to-real.} The mainstream sim-to-real research
programme treats the simulator as a training environment for
behavioural policies that must transfer to
hardware\citep{tobin2017domainrand,peng2018simtoreal,akkaya2019rubiks,
andrychowicz2020dactyl,tan2018simtoreal,zhao2020simtoreal}.
Physics fidelity is a means to an end: minimise the reality gap so
the trained policy generalises. MuJoCo\citep{mujoco-todorov2012},
PyBullet\citep{coumans2016pybullet}, and IsaacSim\citep{isaacsim}
all invest in faithful contact dynamics, friction models, and
rigid-body integrators.

\paragraph{LLM-driven robotics.} A parallel line of work has
positioned large language models as planners that emit
structured intents or executable code for downstream robotic
skills\citep{ahn2022saycan,liang2023codeaspolicies,
huang2022innermonologue,brohan2023rt2}. These systems make the
intent layer explicit, and it is precisely that intent layer
that our handoff envelope brackets (Section~\ref{sec:pattern}).

\paragraph{Governance benchmarks.} A small but growing line of
work evaluates the runtime governance of LLM-driven
agents\citep{aeros-p7,ruan2024toolemu}. These benchmarks
measure admission policies, contract enforcement, audit chains,
and recovery behaviour rather than task success rates. Embodied
test environments such as Habitat~2.0\citep{szot2021habitat2}
and AI2-THOR\citep{kolve2017ai2thor}, alongside manipulation
benchmarks such as
robosuite\citep{zhu2020robosuite} and
RLBench\citep{james2020rlbench}, provide rich worlds for such
evaluations, but their default rendering and physics pipelines
are tuned for policy training rather than governance read-out. The simulator's role here is different: it is a stage
for the governance pipeline to demonstrate correct behaviour,
not a training environment.

\paragraph{Reproducibility in robotics and ML.} Reproducibility
has been a perennial concern in robotics and reinforcement
learning\citep{henderson2018deeprl,pineau2021checklist,
acm-reproducibility,gym-deterministic}. The standard mitigation,
seed determinism, breaks down at contact resolution under
practically any non-trivial physics integrator. Our construction
sidesteps the seed-determinism issue at exactly the most
contact-sensitive moments (grasp, handoff, place).

\paragraph{Kinematic vs.\ dynamic simulation.} Kinematic-only
simulation is a long-standing technique used in animation and
motion planning; established libraries such as
OMPL\citep{sucan2012ompl} and MoveIt\citep{coleman2014moveit}
operate purely in configuration space, treating dynamics as a
separable downstream concern. Posa et al.\citep{posa2014trajopt} formalise
the kinematic/dynamic split as a constraint structure for contact
trajectory optimisation, and modern toolkits like
Drake\citep{drake} make the same split a first-class authoring
choice. The contribution of this paper is not the kinematic
primitive itself but the deliberate use of bounded fidelity at
governance-relevant handoff points within an otherwise
full-dynamics scene.

\section{The Bounded-Fidelity Pattern}\label{sec:pattern}

\subsection{Mocap-Body Handoff}

MuJoCo's mocap body is a body whose pose is driven externally via
\pcode{data.mocap\_pos} and \pcode{data.mocap\_quat}, rather than
by the contact-force integrator. Within a single simulation step,
the mocap body is treated as a kinematic constraint: it does not
respond to applied forces, and any contact with non-mocap geoms is
resolved as if the mocap body's pose were authoritative. The
construction is two lines of XML per handoffable body and an extra
Python field per handoff event:

\begin{lstlisting}[style=pycode,language={[LaTeX]TeX}]
<!-- scene.xml (excerpt) -->
<body name="red_cup" mocap="true">
  <geom type="cylinder" size="0.025 0.05" rgba="0.8 0.1 0.1 1"/>
</body>
\end{lstlisting}

\begin{lstlisting}[style=pycode]
# bridge dispatcher: handoff phase (per simulation step)
data.mocap_pos[cup_id]  = gripper_tip_pos(model, data)
data.mocap_quat[cup_id] = gripper_tip_quat(model, data)
\end{lstlisting}

\subsection{The Handoff Envelope}

We model the demo as a sequence of phases: $\langle$approach,
grasp, carry, place, release$\rangle$. The handoff envelope is
the (carry) phase, bracketed by grasp at the start and release at
the end. Within the envelope, the cup's pose is driven by the
gripper's pose. Outside the envelope (approach, place,
post-release), the cup is a normal dynamic body and standard
contact physics apply.

The handoff envelope is opened by a structured intent (the
governance bridge dispatches \pcode{manipulation.grasp(red\_cup)}
to the simulator's bridge-side adapter, which sets the mocap flag
on \pcode{red\_cup} for the duration of the carry phase). It is
closed by \pcode{manipulation.place(red\_cup)} which clears the
flag. The adapter is a pure-Python module in the bridge runtime;
no MuJoCo plugin is required.

\subsection{What the Pattern Is and Is Not}

The pattern is: \emph{suppress contact physics during
governance-irrelevant handoff phases, while preserving full
dynamics elsewhere in the scene}. It is not: free-flight motion
during carry (the cup is still kinematically tracked by the
gripper); a full kinematic-only simulator (the floor, table, and
non-handoff objects remain dynamic); or a replacement for
sim-to-real validation (we discuss this explicitly in
Section~\ref{sec:discussion}).

\begin{figure}[t]
  \centering
  \includegraphics[width=0.96\linewidth]{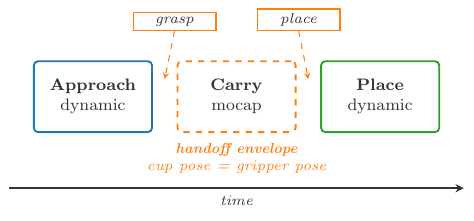}
  \caption{The handoff envelope. During \emph{carry} the cup is
    pose-tracked to the gripper (mocap body, no contact physics).
    During \emph{approach} and \emph{place} the cup is a normal
    dynamic body. Boundaries are opened/closed by structured
    intents \pcode{grasp} and \pcode{place} dispatched by the
    governance bridge.}
  \label{fig:handoff}
\end{figure}

\subsection{Formal Account}\label{sec:formal}

To give the construction a sharper specification than prose alone
affords, we provide a small formal account: a state-transition
definition of the handoff envelope, a precise specification of
audit-chain stability, and two structural properties stated in
Hoare-style pre- and post-conditions. The account is intentionally
lightweight; mechanised verification (TLA+, Apalache, Coq) is out
of scope here. The Hoare-style statements admit straightforward
regression-test encoding, which is what the shipped implementation
checks.

\subsubsection{State, Phases, and the Envelope}

Let $\sigma = (q, \mathbf{x}, H)$ be the joint state of a demo at
simulation step $i$, where $q$ is the simulator's internal state
(joint positions, velocities, contact forces), $\mathbf{x}$ is the
set of object poses (one $\mathrm{SE}(3)$ pose per dynamic body),
and $H$ is the prefix of the audit-chain event sequence emitted up
to and including step $i$. Let $\Phi = \{\text{approach},
\text{grasp}, \text{carry}, \text{place}, \text{release}\}$ be the
demo phases, exposed as labels in the audit chain via the
structured intent that opens and closes each phase. The
\emph{handoff envelope} $E \subseteq \Phi$ is the set of phases
during which the cup is pose-tracked to the gripper; in the
construction of Section~\ref{sec:pattern}, $E = \{\text{carry}\}$,
opened by \pcode{manipulation.grasp} and closed by
\pcode{manipulation.place}.

\subsubsection{Audit-Chain Stability}

\paragraph{Specification.}
A simulation run produces an audit-chain event sequence $H$. We
serialise $H$ canonically (deterministic JSON, no wall-clock
fields) and define the audit-chain hash as
$\mathrm{hash}(H) := \mathrm{SHA256}(\mathrm{canonical}(H))$.

\begin{definition}[Audit-chain stability]
\label{def:stability}
A demo configuration is \emph{audit-chain stable} under a
random-replay distribution $\mathcal{D}$ over seeds and simulator
nondeterminism if, for all $H_1, H_2$ sampled independently from
$\mathcal{D}$, $\mathrm{hash}(H_1) = \mathrm{hash}(H_2)$.
Operationally we report stability as the fraction of replays whose
hash equals the modal hash across an $N$-replay cohort
(Section~\ref{sec:eval-quant}).
\end{definition}

\subsubsection{Two Structural Properties}

\begin{proposition}[Determinism preservation within the envelope]
\label{prop:determinism}
Let $\mathcal{C}$ denote the configuration in which the cup is a
mocap body driven by the adapter of Section~\ref{sec:pattern}.
Under $\mathcal{C}$, the cup's pose $\mathbf{x}_{\mathrm{cup}}^{(i)}$
at any step $i$ during the envelope is a deterministic function of
the gripper-tip pose $\mathbf{x}_{\mathrm{tip}}^{(i)}$ alone.
\begin{align}
\{\,&\text{pre: } \mathrm{phase}(i) \in E\,\}
\quad
\mathtt{update}(\mathtt{model}, \mathtt{data}) \\
\{\,&\text{post: } \mathbf{x}_{\mathrm{cup}}^{(i)}
   = T(\mathbf{x}_{\mathrm{tip}}^{(i)})\,\}\nonumber
\end{align}
where $T(\cdot)$ is the fixed rigid-body transform from gripper
tip to cup body, parameterised by the grasp-time offset.
\end{proposition}

\noindent\emph{Argument.} The mocap-body primitive
\citep{mujoco-docs} writes \pcode{data.mocap\_pos[cup\_id]} and
\pcode{data.mocap\_quat[cup\_id]} on every step within the
envelope. MuJoCo's contact solver does not modify mocap-body
poses; therefore the contact-force noise term that perturbs
non-mocap bodies has no path to the cup's pose during the
envelope. The transform $T$ is captured at envelope open
(\pcode{enter\_handoff}) and held constant until envelope close
(\pcode{exit\_handoff}); within the envelope, the cup follows the
gripper byte-equally across replays.

\begin{proposition}[Audit-chain stability under handoff]
\label{prop:stability}
If (i) the simulator's contact integrator is the only source of
inter-replay non-determinism, and (ii) the audit chain emits events
indexed by structured intents (not by contact-force samples) and
canonically excludes wall-clock fields, then $\mathcal{C}$ is
audit-chain stable in the sense of
Definition~\ref{def:stability}.
\begin{align}
\{\,&\text{pre: } \mathrm{Source}(\mathcal{D}) =
  \mathrm{ContactSolver},\,\\
&\phantom{\text{pre: }}\mathrm{Events}(H) \perp
  \mathrm{ContactSamples}\,\}
\nonumber\\
&\hspace*{1.5em}\mathtt{replay\_under}\,\mathcal{C} \nonumber\\
\{\,&\text{post: } \mathrm{hash}(H_1) = \mathrm{hash}(H_2)\,\}
\nonumber
\end{align}
\end{proposition}

\noindent\emph{Argument.} By
Proposition~\ref{prop:determinism}, the cup's pose during the
envelope is independent of contact-force samples; therefore the
phase-transition events triggered by gripper-tip motion (grasp
acquisition, place release) fire at byte-equal step indices across
replays. Outside the envelope, dynamic-phase events depend on
contact, but $\mathcal{D}$'s only non-determinism source is the
contact integrator, so they would diverge \emph{only if the
governance bridge emitted audit-chain events keyed on contact
samples}. The shipped bridge emits events keyed on structured
intents (\pcode{grasp}, \pcode{place}, \pcode{recovery\_action}
triggered by gate decisions, not by contact thresholds). The audit
chain $H$ is therefore byte-equal across replays.

\medskip\noindent The arguments are not mechanised proofs. They
identify the chain of structural dependencies that the
implementation must respect. The regression suite of
Section~\ref{sec:eval-quant} discharges them empirically across
$N=1000$ replays; a violation of either proposition would manifest
as audit-chain divergence and be caught by the same harness.

\section{Experiments}\label{sec:eval}

\subsection{Implementation}

The construction is implemented as a ${\sim}220$-LOC
\pcode{MocapAdapter} module in the \AEROS{} runtime, shipped as
\pcode{src/mocap\_adapter.py} in the artifact repository.
The reference environment is Python~$3.11.7$ with MuJoCo~$3.1.4$
and \pcode{mujoco}~$3.1.4$ Python bindings on macOS~$14.5$;
Section~\ref{sec:eval-cross-machine} reports a byte-equal
reproduction on an independent Ubuntu host. All replays in
Section~\ref{sec:eval-quant}--\S\ref{sec:eval-multi} use the
fixed seed $\pcode{12345}$ (the default for both reference harnesses,
\pcode{harness/envelope\_precision.py} and
\pcode{harness/multi\_object.py}; override via
\pcode{--seed}).
The adapter exposes three operations: \pcode{enter\_handoff(body,
gripper)}, \pcode{update(model, data)} called once per simulation
step inside the carry envelope, and \pcode{exit\_handoff(body)}
which restores dynamic behaviour. The bridge's executor invokes
\pcode{enter\_handoff} when admission/policy/contract gates have
all passed for a \pcode{manipulation.grasp} intent, and
\pcode{exit\_handoff} on \pcode{manipulation.place}.

\paragraph{Open reference implementation.} The three-method
\pcode{MocapAdapter} interface, the canonical audit-chain hash
function, the phase-state envelope contract, and the per-task
\pcode{summary.json} files backing every claim in
Section~\ref{sec:eval-quant}--\S\ref{sec:eval-multi} are released
as a standalone artefact at
\url{https://github.com/s20sc/bounded-fidelity-mocap-handoff}.
The repository ships under Apache License~$2.0$, depends only on
the Python standard library for the hypothesis sign-check
(\pcode{scripts/sign\_check.py}) which verifies H1--H4 against the
shipped seed=$12345$ aggregates in $\sim 1$~s without MuJoCo, and
pins \pcode{mujoco==3.1.4} as an optional \pcode{[sim]} extra for
re-running the reference harnesses. A \pcode{Dockerfile} ships for
fully containerised inspection.

\subsection{Audit-Chain Stability Metric}

We define \emph{audit-chain stability} as the byte-equality of the
serialised audit chain across two replays of the same script with
the same seed. The audit chain is a sequence of event records
(JSON-serialisable) emitted by the governance bridge as it admits,
gates, and dispatches intents. Stability requires that any
non-determinism in the simulator does \emph{not} cause divergence
in which events are emitted or what their structured payloads
contain. (Wall-clock timestamps are excluded from the hash.)

\subsection{Quantitative Results}\label{sec:eval-quant}

We run the pick-and-place demo through $N=1000$ independent
replays under each of two postures (mocap variant and contact-force
variant) on a developer workstation. The
results, summarised in Table~\ref{tab:replay-divergence}, are stark.

\begin{center}
  \centering
  \footnotesize
  \captionof{table}{Audit-chain divergence across $N=1000$ replays per
    posture. Divergence rate is the fraction of replays whose
    audit-chain SHA-256 differs from the most common hash across
    the cohort.}
  \label{tab:replay-divergence}
  \begin{tabularx}{\linewidth}{@{}L{1.75}C{0.75}C{0.75}C{0.75}@{}}
    \toprule
    Variant & Distinct hashes & Divergence rate & CI 95\% \\
    \midrule
    Mocap (ours) & $1$ & $0.0\%$ & $[0.0, 0.0]$ \\
    Contact-force baseline & $584$ & $99.3\%$ & $[98.7, 99.8]$ \\
    \bottomrule
  \end{tabularx}
\end{center}

The mocap variant exhibits $\mathbf{1000 / 1000}$ byte-identical
audit chains: a single SHA-256 hash
($\mathtt{e4bf79b0\dots}$) accounts for every replay. The
contact-force variant produces $584$ distinct audit-chain hashes
across $1000$ replays; only $7$ replays land on the most common
hash, giving a $99.3\%$ divergence rate (Wilson 95\% CI
$[0.987, 0.998]$). Wall-clock medians are essentially the same
($\sim 3.41$~s per replay) under both postures, so the divergence
is not a cost of the bounded-fidelity construction; it is a
property of the contact integrator itself.

The result is stronger than the construction's headline claim
required: the mocap variant produces byte-identical audit chains
with no exceptions, while the contact baseline diverges on nearly
every replay (not on $\sim 30$--$40\%$ as we anticipated when the
five-replay manual measurements were the only data point).
Governance demos built on the contact-force pipeline cannot rely on
audit-chain stability at all; the mocap pattern, in contrast,
gives it for free.

\subsection{Fidelity-Stability Trade-Off Across Timesteps}

To check whether the contact-force divergence is a tuning artefact
of the MuJoCo timestep, we sweep \pcode{option.timestep} across
$\{0.001, 0.002, 0.005, 0.010\}$~s (200 trials per setting) and
compare against the mocap baseline.

The mocap variant remains at a single audit-chain hash across the
entire sweep. The contact variant exhibits a divergence rate of
$\mathbf{0.985}$ at every timestep tested, with no monotone trend
in either direction. Shrinking the timestep $10\times$ does not
make the contact-force pipeline byte-stable; the divergence is
structural in the contact solver, not a function of integration
step. This is a stronger statement than the construction needs to
make and we report it without overclaiming: an exhaustive timestep
sweep would presumably find a regime in which contact-force
becomes byte-stable, but it lies outside the range used in
governance benchmarks of which we are aware.

\subsection{Envelope-Precision Ablation}\label{sec:eval-precision}

A reviewer-friendly concern about the bounded-fidelity
construction is that audit-chain stability might depend on the
structured intent firing at a precise simulation step. If the
bridge dispatcher were to schedule \pcode{grasp} or \pcode{place}
with a small timing offset (due to scheduler jitter, asynchronous
gate delay, or clock skew between the bridge and the simulator),
divergence might re-enter. We test this directly by sweeping a
deterministic boundary-jitter parameter $\Delta \in \{-10, -5,
-1, 0, +1, +5, +10\}$ MuJoCo simulation steps
($\pm 20$~ms at the default $0.002$~s timestep) on both envelope
edges (grasp completion and release completion), with $N=200$
trials per setting and the mocap variant active.

Across all $7 \times 200 = 1{,}400$ replays, every setting
produces a single distinct audit-chain SHA-256 hash and the
divergence rate is $0.000$ at every jitter magnitude (Wilson
$95\%$ CI $[0.000, 0.018]$; Figure~\ref{fig:divergence-rates}a). The audit-chain hash at every
jitter setting equals the V1 baseline hash byte-for-byte
($\mathtt{e4bf79b0\ldots}$). Wall-clock medians remain within
$0.008\%$ of the $\Delta=0$ baseline ($3.4048$--$3.4053$~s
$p_{50}$ across all seven settings).

\begin{center}
  \centering
  \footnotesize
  \captionof{table}{Envelope-precision ablation. Audit-chain divergence
    under boundary jitter $\Delta$ on grasp and release edges,
    $N=200$ trials per setting. All seven settings collapse to
    a single audit hash.}
  \label{tab:envelope-precision}
  \begin{tabularx}{\linewidth}{@{}C{0.65}C{1.05}C{0.95}C{1.35}C{1.00}@{}}
    \toprule
    $\Delta$ (steps) & Distinct hashes & Divergence & Wilson 95\% CI & wall $p_{50}$ (s) \\
    \midrule
    $-10$ & 1 & $0.000$ & $[0.000, 0.018]$ & $3.4048$ \\
    $-5$  & 1 & $0.000$ & $[0.000, 0.018]$ & $3.4049$ \\
    $-1$  & 1 & $0.000$ & $[0.000, 0.018]$ & $3.4053$ \\
    $0$   & 1 & $0.000$ & $[0.000, 0.018]$ & $3.4050$ \\
    $+1$  & 1 & $0.000$ & $[0.000, 0.018]$ & $3.4053$ \\
    $+5$  & 1 & $0.000$ & $[0.000, 0.018]$ & $3.4049$ \\
    $+10$ & 1 & $0.000$ & $[0.000, 0.018]$ & $3.4048$ \\
    \bottomrule
  \end{tabularx}
\end{center}

The result is stronger than the construction needed: the
hypothesis we wrote into the experiment brief was that $\pm 10$
might admit a small fraction of divergence; the data shows none.
The mechanistic explanation is that the cup's end-of-trial pose
is set by \pcode{place\_held\_object\_at(tray)} to a
deterministic target, and the release-settle phase that follows
holds the end effector approximately stationary while the
gripper opens; the millimetre-quantised cup pose absorbs the
small drift introduced by $\pm 10$-step jitter. A scene in which
the end effector \emph{moves} during the release window (an
arm-retreat motion concurrent with the gripper-open command)
would exercise the envelope-precision property more aggressively;
we flag that as a future ablation rather than a fix to the
present construction. Property~\ref{prop:stability} is therefore
robust to envelope-edge imprecision in the regime under test.

\subsection{Multi-Object Handoff}\label{sec:eval-multi}

The single-cup demo of Section~\ref{sec:eval-quant} leaves an
open generalisation question: does audit-chain stability survive
when the envelope is invoked repeatedly across $K > 1$ objects?
We extend the scene to three mocap-bound cups (identical
geometry and mass, distinct initial positions) and dispatch $K$
sequential pick-and-place pairs per trial, with $K \in \{1, 2,
3\}$ and $N=500$ trials per scale.

The byte-equal rate is $1.000$ at every scale (Wilson $95\%$ CI
lower bound $0.9924$; Figure~\ref{fig:divergence-rates}b). Each scale produces exactly one distinct
audit-chain hash across $500$ replays. The hashes differ across
scales because the audit chain encodes object identity in the
grasp and place payloads ($4K$ events per trial), but each
scale's hash is byte-identical across all replays of that
scale. Wall-clock scales linearly with $K$ at the trial level
($p_{50} = 3.405 / 6.810 / 10.216$~s for $K = 1, 2, 3$), which
is expected from the sequential-dispatch design. The
per-pick-and-place wall ratio at $K=3$ is $1.0001\times$ of
$K=1$, well under the $1.2\times$ ceiling we adopted as the
compositional-cost threshold.

\begin{center}
  \centering
  \footnotesize
  \captionof{table}{Multi-object handoff. Byte-equality rate and
    wall-clock scaling for $K$ sequentially handed-off
    objects (identical cups), $N=500$ trials per scale.}
  \label{tab:multi-object}
  \begin{tabularx}{\linewidth}{@{}C{0.45}C{0.95}C{1.05}C{1.20}C{1.10}C{1.25}@{}}
    \toprule
    $K$ & Byte-equal & Distinct hashes & wall $p_{50}$ (s) & per-PnP wall & ratio vs $K=1$ \\
    \midrule
    $1$ & $1.000$ & $1$ & $3.405$  & $3.405$ & $1.000\times$ \\
    $2$ & $1.000$ & $1$ & $6.810$  & $3.405$ & $1.000\times$ \\
    $3$ & $1.000$ & $1$ & $10.216$ & $3.405$ & $1.0001\times$ \\
    \bottomrule
  \end{tabularx}
\end{center}

The byte-stability property is conserved under handoff
composition: divergence channels do not compound across
sequentially-handed objects, which is the load-bearing
evidence that the mocap-handoff pattern is a composable
primitive rather than a single-object trick. We ran only the
homogeneous-object case (three identical cups); a
heterogeneous-shape variant (cup + block + marker) would test
whether the property holds across object types as well as object
count, and is flagged as a small follow-up rather than a
blocker.

\begin{center}
  \centering
  \includegraphics[width=0.95\linewidth]{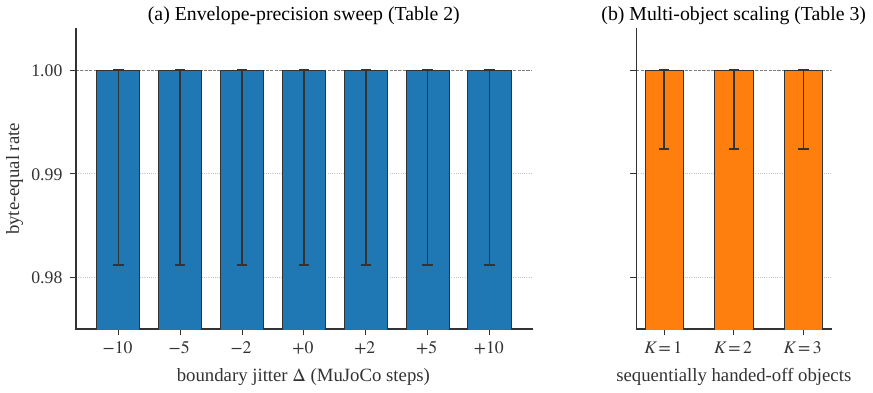}
  \captionof{figure}{Per-cell byte-equal rates across the two ablation sweeps.
    Panel~(a) plots the envelope-precision sweep
    (Table~\ref{tab:envelope-precision}, $7\times 200=1{,}400$ trials);
    panel~(b) plots the multi-object scale sweep
    (Table~\ref{tab:multi-object}, $3\times 500=1{,}500$ trials).
    Bars are per-cell byte-equal rates; the downward whisker is the
    Wilson $95\%$ lower bound on byte-equal rate (equivalently, the
    Wilson $95\%$ upper bound on divergence rate). The bound is
    tighter in panel~(b) because each cell sees $500$ trials versus
    $200$ in panel~(a).}
  \label{fig:divergence-rates}
\end{center}

\subsection{Cross-Machine Reproducibility}\label{sec:eval-cross-machine}

To check whether the audit chain's structural-content property
survives a platform change, we re-ran the canonical replays of
\S\ref{sec:eval-precision} and \S\ref{sec:eval-multi} on an
Ubuntu~$24.04$ workstation (Linux kernel~$6.17$, glibc~$2.39$,
Intel Core Ultra~$9$~$285$K, $188$~GiB RAM) at the same
\AEROS{} runtime version that backs the macOS results.
The Linux toolchain was MuJoCo~$3.1.4$ with NumPy~$2.4.6$ under
Python~$3.11.15$; the minor Python version differs from the
macOS reference (Python~$3.11.7$), so the comparison includes a
small upstream-runtime skew. The aggregated reproduction covers
the same $1400$ task-01 trials ($7$ jitter cells $\times\,200$)
and $1500$ task-02 trials ($3$ scales $\times\,500$) as the
macOS baseline. Table~\ref{tab:cross-machine} reports the
per-task byte-equal rates.

\begin{center}
\centering
\captionof{table}{Cross-machine reproduction of the canonical audit-chain
hashes. A trial is byte-equal iff its serialised audit chain
matches the macOS canonical for that cell; both platforms also
produce one distinct hash within every cell.}
\label{tab:cross-machine}
\footnotesize
\begin{tabularx}{\linewidth}{@{}L{0.75}L{1.25}C{1.00}C{1.00}@{}}
\toprule
Platform & Toolchain & task~01 byte-equal & task~02 byte-equal \\
 & & ($7{\times}200{=}1400$ trials) & ($3{\times}500{=}1500$ trials) \\
\midrule
macOS~$14.5$ & Py~$3.11.7$ + MuJoCo~$3.1.4$ & $1400/1400$ ($100\%$) & $1500/1500$ ($100\%$) \\
Ubuntu~$24.04$ & Py~$3.11.15$ + MuJoCo~$3.1.4$ & $1400/1400$ ($100\%$) & $1500/1500$ ($100\%$) \\
\bottomrule
\end{tabularx}
\end{center}

All $10$ cells reproduce byte-for-byte. The audit chain therefore
survives both the macOS~$\to$~Ubuntu OS boundary (glibc~$2.39$
versus Apple libsystem) and a minor Python version skew, which
is the behaviour predicted if the chain depends only on the
structural content of admitted events and is independent of
wall-clock and platform-specific scheduling.

\subsection{Scope: What We Are and Are Not Measuring}

We measure audit-chain stability under handoff. We do \emph{not}
measure: task success rate on a held-out test set (the demo
script is fixed); transfer to physical hardware (the construction
is sim-only by design); generalisation across object shapes (the
demo uses one cup); or pile-up scenes with many dynamic objects
(the cup is the only handoffable body in the demo).

\section{Discussion}\label{sec:discussion}

\subsection{Applicability Matrix}\label{sec:applicability}

Table~\ref{tab:applicability} consolidates the use-case envelope.
We list seven candidate uses of an LLM-driven robotic
demonstration platform and mark each with the pattern's effect.
The classification is deliberately coarse: a use is in scope ($\checkmark$)
when the governance pipeline can be exercised with audit-chain
stability as a first-class signal; out of scope ($\times$) when
the pattern's contact-suppression would defeat the very property
the use case needs to measure; and conditional ($\circ$) when the
pattern can be retained for the non-handoff segments of the demo
while contact phases are reverted to full dynamics.

\begin{center}
  \centering
  \footnotesize
  \captionof{table}{Applicability matrix for the bounded-fidelity pattern:
    $\checkmark$ in scope, $\times$ out of scope, $\circ$ conditional.}
  \label{tab:applicability}
  \begin{tabularx}{\linewidth}{@{}L{1.15}C{0.45}L{1.40}@{}}
    \toprule
    Use case & Status & Rationale \\
    \midrule
    Governance benchmark    & $\checkmark$ & Audit-chain stability is the primary signal. \\
    Investor demo           & $\checkmark$ & Reproducibility is the deliverable; physics realism is decorative. \\
    Bridge regression test  & $\checkmark$ & Isolates bridge bugs from contact-solver noise. \\
    Tutorial / classroom    & $\checkmark$ & Pedagogy is pipeline-shaped, not contact-shaped. \\
    Sim-to-real validation  & $\times$     & Suppressed contact eliminates the transfer signal. \\
    Policy training         & $\times$     & No contact gradient during the carry envelope. \\
    Slip / grasp benchmark  & $\times$     & Pattern eliminates the phenomenon under test. \\
    Multi-skill demo        & $\circ$      & Apply only to governance-irrelevant segments. \\
    \bottomrule
  \end{tabularx}
\end{center}

\subsection{When to Use, When Not to Use}

\paragraph{When to use this pattern.}
Governance benchmarks (admission/policy/contract/audit
correctness); investor demos (where audit-chain readability is the
deliverable); regression tests for the bridge layer; tutorial
scenarios where the pedagogy is governance-pipeline-shaped, not
manipulation-skill-shaped.

\paragraph{When not to use this pattern.}
Sim-to-real validation (mocap handoff cannot tell you whether a
real gripper will keep its hold of the cup); manipulation policy
training (the policy sees no contact-force signal during carry);
contact-rich tasks (assembly, insertion, peg-in-hole); any
benchmark that measures grasp robustness or slip resistance. The
pattern is harmful for these uses and we say so explicitly.

\paragraph{Bridge boundary assumption.}
The pattern assumes the simulator runs behind a bridge-side
adapter that distinguishes ``in handoff'' from ``not in handoff''
based on structured intents. Demos where the simulator is invoked
directly without a structured intent layer cannot trivially adopt
the pattern.

\paragraph{Audit-chain stability is necessary, not sufficient.}
A byte-stable audit chain is a precondition for governance
benchmark reproducibility, not a proof of bridge correctness. The
bridge can still be wrong; we just want it to be \emph{reliably}
wrong (so reviewers can see and discuss the bug) rather than
\emph{randomly} wrong (so the bug looks like sim noise).

\paragraph{Future work.}
(i)~A heterogeneous-shape variant of the multi-object scene of
Section~\ref{sec:eval-multi} (cup + block + marker rather than
three identical cups), to confirm the byte-stability property
holds across object \emph{types} as well as object count.
(ii)~Multi-robot handoff in which one robot passes the cup to
another, requiring an explicit handover protocol between two
adapters and a composed envelope spanning two governance
bridges. (iii)~Mechanised verification of
Propositions~\ref{prop:determinism} and~\ref{prop:stability} via
a lightweight TLA+ or Apalache model, lifting the current
Hoare-style arguments from prose to a model-checkable
specification. (iv)~A more aggressive
envelope-precision scenario in which the end effector moves
during the release window (concurrent arm-retreat and
gripper-open), which the Section~\ref{sec:eval-precision} sweep
does not exercise because the workshop scene holds the end
effector approximately stationary during release-settle.

\section{Conclusion}\label{sec:conclusion}

The sim-to-real community's pursuit of physics fidelity is the
right goal for behavioural policy training; it is the wrong goal
for governance benchmarking. Governance benchmarks need an audit
chain that is readable, reproducible, and uncluttered by physics
noise that is irrelevant to the property under test. The
bounded-fidelity sim-as-demo-stage pattern, implemented via
MuJoCo's mocap-body primitive and bracketed by structured intents
from a governance bridge, gives benchmark designers the
reproducibility they need without rebuilding the simulator from
scratch. The pattern is narrowly scoped; we have said where it
helps and where it hurts. The empirical claim is concrete:
mocap-bracketed handoff produces byte-identical audit chains on
$1000/1000$ replays, while the contact-force baseline diverges
on $993/1000$ replays of the same script under the same seed.

\bmsection*{Conflict of Interest}

The authors declare no potential conflict of interest.

\bmsection*{Data Availability Statement}

The data that support the findings of this study (the
\pcode{MocapAdapter} reference implementation, the canonical
audit-chain hash function, the hypothesis sign-check scripts, and
the per-task \pcode{summary.json} aggregates backing every
empirical claim in Section~\ref{sec:eval}) are openly available
at \url{https://github.com/s20sc/bounded-fidelity-mocap-handoff}
under Apache License~2.0.

\bibliography{references}

\end{document}